\documentclass[10pt,twocolumn,letterpaper]{article}

\usepackage{cvpr}              

\usepackage{graphicx}
\usepackage{amsmath}
\usepackage{amssymb}
\usepackage{booktabs}

\usepackage{mathtools}
\usepackage{amsmath}

\usepackage{amssymb}
\usepackage{amsfonts}
\usepackage{amsopn}
\usepackage{graphicx} 
\usepackage{textcomp}
\usepackage{xfrac}
\usepackage{bbm}
\usepackage{overpic}
\usepackage{wrapfig}
\usepackage[ruled,vlined,linesnumbered]{algorithm2e}
\usepackage{multirow}
\usepackage{array}

\usepackage{color}
\definecolor{green}{rgb}{0, 0.5, 0}
\definecolor{orange}{rgb}{0.8, 0.6, 0.2}
\definecolor{red}{rgb}{1.0, 0.0, 0.0}
\definecolor{teal}{rgb}{0.0, 0.4, 0.4}
\definecolor{purple}{rgb}{0.65,0,0.65}
\definecolor{saffron}{rgb}{0.95,0.75,0.2}
\definecolor{turquoise}{rgb}{0.0,0.5,0.5}
\definecolor{brown}{rgb}{0.5, 0.16, 0.16}

\usepackage{overpic}
\usepackage{currfile} 

\newlength\savedwidth

\newcommand{\ys}[1]{{\color{black}#1}}
\newcommand{\revise}[1]{{\color{black}#1}}

\definecolor{lightgray}{rgb}{0.6, 0.6, 0.6}

\usepackage[normalem]{ulem}

\newcommand{\hidecomment}[1]{}

\newcommand{\bX}{\mathbf{X}}

\newcommand{\bc}{\mathbf{c}}
\newcommand{\bF}{\mathbf{F}}
\newcommand{\bV}{\mathbf{V}}

\newcommand{\bW}{\mathbf{W}}
\newcommand{\bz}{\mathbf{z}}
\newcommand{\bh}{\mathbf{h}}
\newcommand{\bQ}{\mathbf{Q}}
\newcommand{\bK}{\mathbf{K}}

\newcommand{\bZ}{\mathbf{Z}}
\newcommand{\bS}{\mathbf{S}}

\usepackage{xspace}

\usepackage[pagebackref,breaklinks,colorlinks]{hyperref}

\usepackage[capitalize]{cleveref}
\crefname{section}{Sec.}{Secs.}
\Crefname{section}{Section}{Sections}
\Crefname{table}{Table}{Tables}
\crefname{table}{Tab.}{Tabs.}

\def\cvprPaperID{4132} 
\def\confName{CVPR}
\def\confYear{2022}

\begin{document}

\title{RayMVSNet: Learning Ray-based 1D Implicit Fields for Accurate Multi-View Stereo}

\author{
    \hfill
	Junhua Xi\thanks{Joint first authors}\qquad
	Yifei Shi\footnotemark[1]\qquad
	Yijie Wang\qquad
    Yulan Guo\qquad
	Kai Xu\thanks{Corresponding author: kevin.kai.xu@gmail.com}
	\hfill
	\vspace{0.1cm}
	\\
	\hfill
	National University of Defense Technology
    \hfill
}

\maketitle

\begin{abstract}
Learning-based multi-view stereo (MVS) has by far centered around 3D convolution on cost volumes. Due to the high computation and memory consumption of 3D CNN, the resolution of output depth is often considerably limited. Different from most existing works dedicated to adaptive refinement of cost volumes, we opt to directly optimize
the depth value along each camera ray, mimicking the range (depth) finding of a laser scanner. This reduces the MVS problem to ray-based depth optimization which is much more light-weight than full cost volume optimization. In particular, we propose RayMVSNet which learns sequential prediction of a 1D implicit field along each camera ray with the zero-crossing point indicating scene depth. This sequential modeling, conducted based on transformer features, essentially learns the epipolar line search in traditional multi-view stereo. We also devise a multi-task learning for better optimization convergence and depth accuracy. Our method ranks top on both the DTU and the Tanks \& Temples datasets over all previous learning-based methods, achieving overall reconstruction score of $0.33$mm on DTU and f-score of $59.48\%$ on Tanks \& Temples.
\end{abstract} 


\section{Introduction}
\label{sec:intro}
Learning-based multi-view stereo has gained a surge of attention recently since the seminal work of MVSNet~\cite{yao2018mvsnet}.
The core idea of MVSNet and many followup works is to construct a 3D cost volume in the frustum of the reference view through warping the image features of several source views onto a set of fronto-parallel sweeping planes at hypothesized depths.
3D convolutions are then conducted on the cost volume to extract 3D geometric features and regress the final depth map of the reference view.

Most existing methods are limited to low-resolution cost volume since 3D CNN is generally computation and memory consuming.
Several recent works proposed to upsample or refine cost volume aiming at increasing the resolution of output depth maps~\cite{gu2020cascade,yang2020cost,cheng2020deep}. Such refinement, however, still needs to trade off between depth and spatial (image) resolutions. For example, CasMVSNet~\cite{gu2020cascade} opts to narrow down the range of depth hypothesis to allow high-res depth estimation, matching the spatial resolution of input RGB. 3D convolution is then naturally confined within the narrow band, thus degrading the efficacy of 3D feature learning.

\begin{figure}[t] \centering
	\begin{overpic}[width=0.97\linewidth,tics=10]{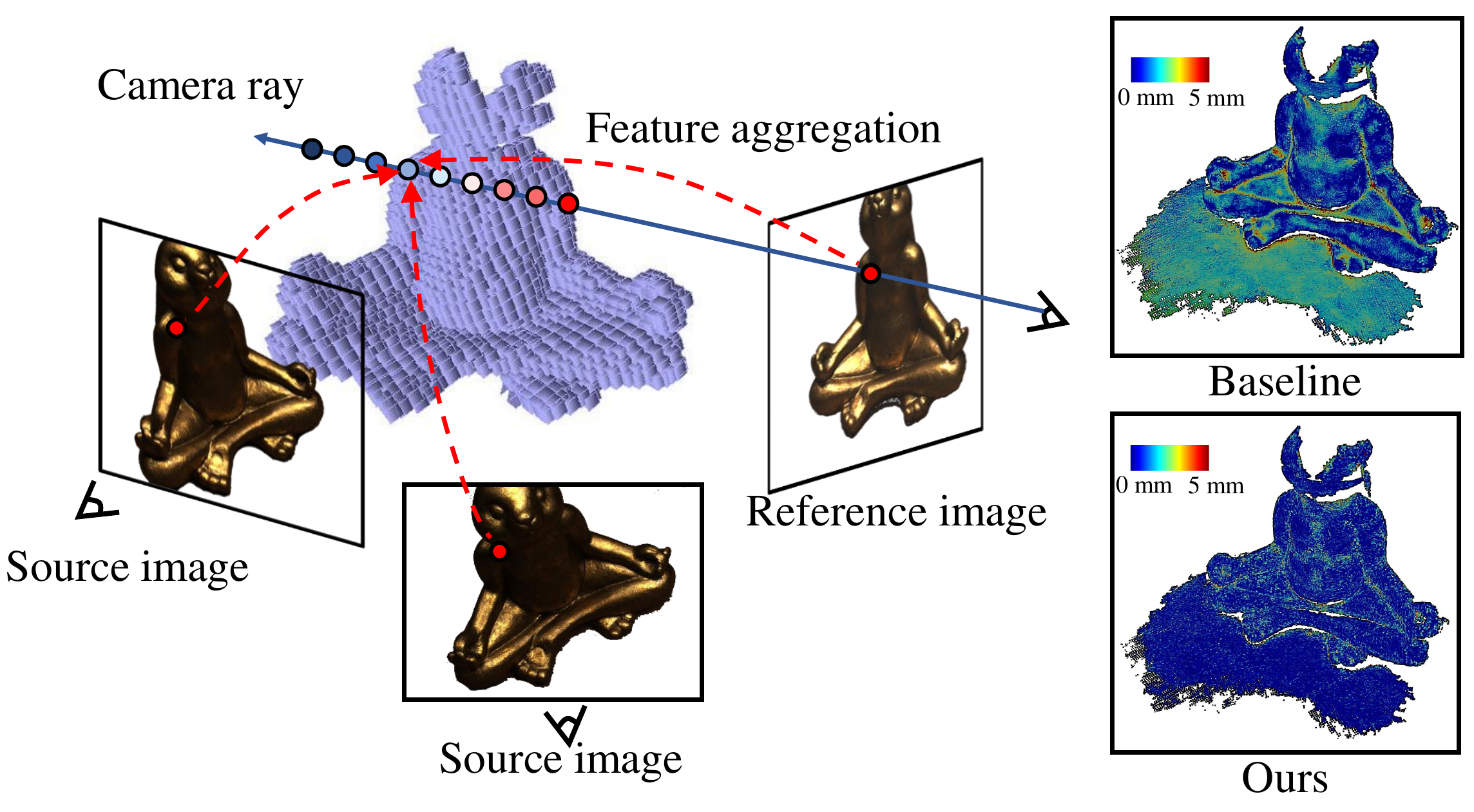}
   \end{overpic}\vspace{-5pt}
   \caption{RayMVSNet performs multi-view stereo via predicting 1D implicit fields on a camera ray basis.
   The sequential prediction of 1D field is light-weight and the monotonicity of ray-based distance field around surface-crossing points facilitates robust learning, leading to more accurate depth estimation than the purely cost-volume-based baselines such as \revise{MVSNet~\cite{yao2018mvsnet}}.
   }
   \label{fig:teaser}\vspace{-5pt}
\end{figure}

In fact, depth map is view-dependent although cost volume is not. Since the target is depth map, refining the cost volume seems neither economic nor necessary. There could be a large portion of the cost volume invisible to the view point. In this work, we advocate direct optimizing the depth value along each camera ray, mimicking the range (depth) finding of a laser scanner. This allows us to convert the MVS problem into ray-based depth optimization which is, individually, a much more light-weight task than full cost volume optimization. We formulate the ``range finding'' of each camera ray as learning a 1D implicit field along the ray whose zero-crossing point indicates the scene depth along that ray \ys{(Figure~\ref{fig:teaser})}. To do so, we propose RayMVSNet which learns sequential modeling of multi-view features along camera rays based on recurrent neural networks.


Technically, we propose two critical designs to facilitate learning accurate ray-based 1D implicit fields. \emph{Firstly}, the sequential prediction of 1D implicit field along a camera ray is essentially conducting an epipolar line search~\cite{andrew2001multiple} for cross-view feature matching whose optimum corresponds to the point of ray-surface intersection. To learn this line search, we propose \emph{Epipolar Transformer}. Given a camera ray of the reference view, it learns the matching correlation of the pixel-wise 2D features of each source view based on attention mechanism. The transformer features of all views, together with (low-res) cost volume features, are then concatenated and fed into an LSTM~\cite{hochreiter1997long} for implicit field regression. Figure~\ref{fig:epipolar} visualizes how epipolar transformer selects reliable matching features from different views.

\emph{Secondly}, we confine the sequential modeling for each camera ray within a fixed-length range centered around the hypothesized surface-crossing point given by the vanilla MVSNet. This makes the output 1D implicit field along each ray monotonous, which is normalized to $[-1,1]$. Such restriction and normalization lead to significant reduction of learning complexity and improvement of result quality. We devise two learning tasks: 1) sequential prediction of signed distance at a sequence of points sampled in the fixed-length range and 2) regression of the zero-crossing position on the ray. A carefully designed loss function correlates the two tasks. Such multi-task learning approach yields highly accurate estimation of per-ray surface-crossing points.

Learning view-dependent implicit fields has been well-exploited in neural radiance fields (NeRF)~\cite{mildenhall2020nerf} with great success. Recently, NeRF was combined with MVSNet for better generality~\cite{chen2021mvsnerf}. Albeit sharing conceptual similarity, our work is a completely different from NeRF. \emph{First}, NeRF (including MVSNeRF~\cite{chen2021mvsnerf}) is designed for novel view synthesis, a different task from MVS. \emph{Second}, the radiance field in NeRF is defined and learned in continuous 3D space and camera rays are used only in the volume rendering stage. In our RayMVSNet, on the other hand, we explicitly learn 1D implicit fields on a camera ray basis.

RayMVSNet ranks top on both the DTU and the Tanks \& Temples datasets over all learning-based methods. It achieves overall reconstruction score of $0.33$mm on DTU, and f-score of $59.48\%$ on Tanks \& Temples.
Notably, since all rays share weights for the LSTM and the epipolar transformer, the RayMVSNet model is light weight.
Moreover, the computation for each ray is highly parallelizable.

Our work makes the following contributions:
\begin{itemize}
  \vspace{-6pt}
  \item A novel formulation of deep MVS as learning ray-based 1D implicit fields.
  \vspace{-8pt}
  \item An epipolar transformer designed to learn cross-view feature correlation with attention mechanism.
  \vspace{-8pt}
  \item A multi-task learning approach to sequential modeling and prediction of 1D implicit fields based on LSTM.
  \vspace{-8pt}
  \item A challenging test set focusing on regions with specular reflection, shadow or occlusion based on the DTU dataset~\cite{aanaes2016large} and associated extensive evaluations.
\end{itemize}

\section{Related Work}
\label{sec:related}

\paragraph{Learning-based MVS.}
Recent advances have made remarkable progress on learning-based MVS. Hartmann et al.~\cite{hartmann2017learned} first propose to learn the multi-patch similarity from two views by a Siamese convolutional network. SurfaceNet~\cite{ji2017surfacenet} and DeepMVS~\cite{huang2018deepmvs} warp the multi-view images into the 3D cost volume and adopt 3D neural networks to estimate the geometry.
MVSNet~\cite{yao2018mvsnet} proposes a differentiable homography and leverages 3D cost volume in a learning pipeline.
MVSNet aggregates contextual information by a 3D convolutional network.
However, the high computation and memory consumption restrict the output depth resolution, limiting its scalability in large scenes.

To reduce the requirements, many follow-up works have been developed.
R-MVSNet~\cite{yao2019recurrent} proposes to regularize the 2D cost maps along the depth direction so the memory consumption could be greatly reduced.
Point-MVSNet~\cite{chen2019point} first computes the coarse depth with a low-resolution cost volume and then uses a point-based refinement network to generate the high-resolution depth map.
CasMVSNet~\cite{gu2020cascade} adopts a cascade cost volume to gradually narrow the depth range and increase the cost volume resolution.
Similar ideas are later explored to reduce the memory cost of 3D convolutions and/or increase the depth quality, \ys{such as coarse-to-fine depth optimization~\cite{yang2020cost,yu2020fast,yan2020dense,xu2020learning,ma2021epp,cheng2020deep,xu2021digging}, attention-based feature aggregation~\cite{luo2020attention,yu2021attention,zhang2021long,wei2021aa}, and patch matching-based method~\cite{luo2019p,wang2021patchmatchnet}.}
Unlike these works, RayMVSNet optimizes the depth on each camera viewing ray instead of the 3D volume, which is more light-weight.

\revise{Multi-view feature aggregation is one of the most crucial components in learning-based MVS. Previous works adopted various solutions to learn mutual correlations~\cite{zhao2018triangle}, avoiding the influences of incorrect matches caused by occlusion. Popular solutions include the visibility-based aggregation~\cite{zhang2020visibility,chen2020visibility}, the attention-based aggregation~\cite{yi2020pyramid,wei2021aa}, etc.
RayMVSNet follows the attention-based aggregation route. Nevertheless, it learns feature aggregation at each 3D point, instead of the entire image or volume, thus greatly reducing the memory consumption.}


\begin{figure*}[t] \centering
	\begin{overpic}[width=0.97\linewidth,tics=10]{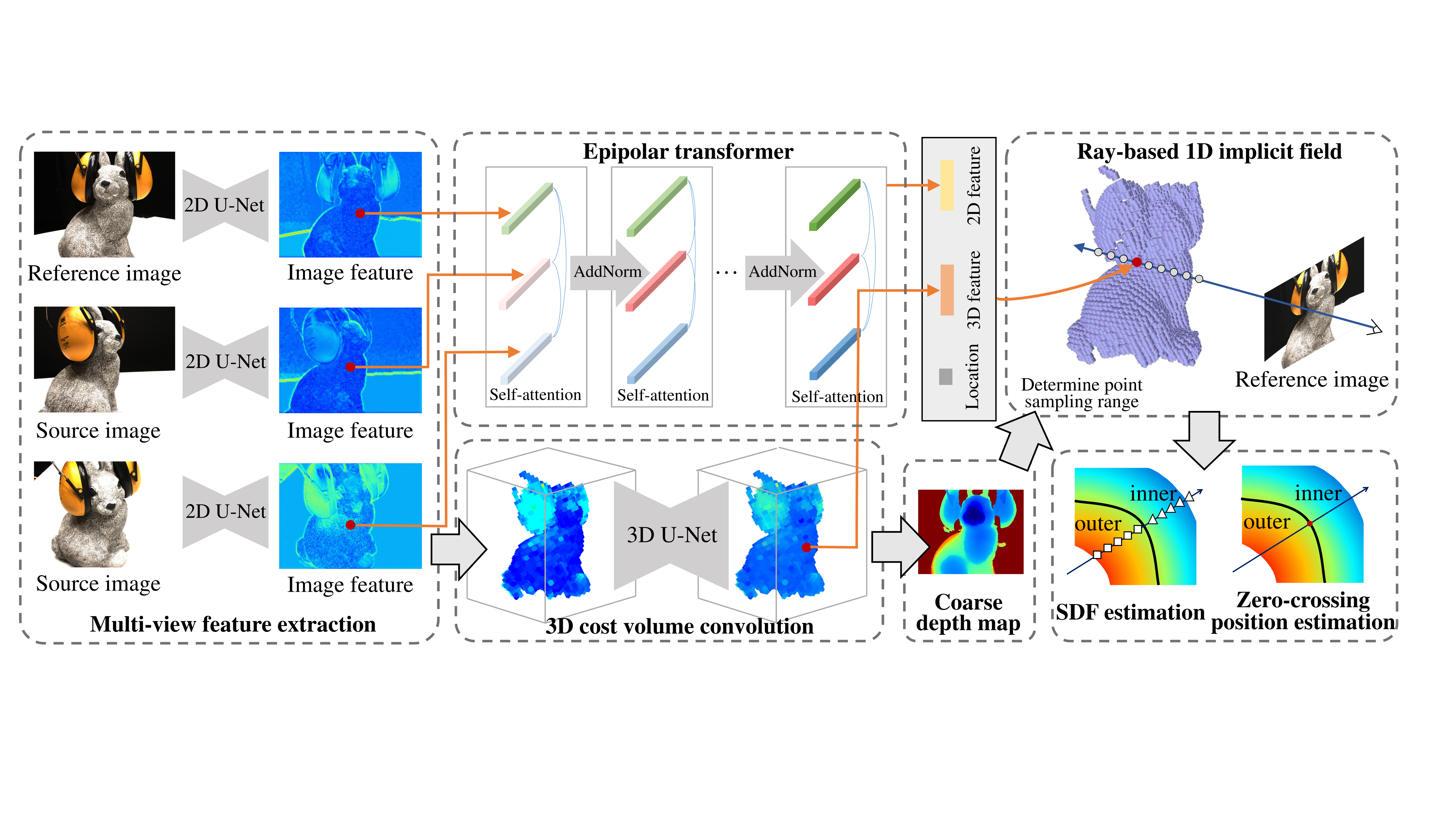}
   \end{overpic}\vspace{-5pt}
   \caption{Method overview. Given multiple overlapping RGB images, the multi-view image features are extracted by a 2D U-Net.
   The coarse depth map is then estimated by a coarse 3D cost volume. 2D multi-view image features are then correlated and aggregated by epipolar transformer.
   At last, the 1D implicit field is learnt on each camera viewing ray to simultaneously estimate the SDF of the sampled points and the location of the zero-crossing point.
   }
   \label{fig:overview}\vspace{-5pt}
\end{figure*} 

\vspace{-8pt}
\paragraph{Learning Implicit Representation.}
Many works have attempted learning shape representation based on implicit fields.
Implicit field shows promising results on facilitating a variety number of problems, such as shape reconstruction~\cite{zhang2021learning,zheng2021deep,nelson2020learning,deng2021deformed} and rendering~\cite{mildenhall2020nerf,takikawa2021neural}.
DeepSDF~\cite{park2019deepsdf} proposes to predict the magnitude of 3D point to indicate the distance to the surface boundary and a sign to determine whether the point is inside or outside of the shape.
IM-Net~\cite{chen2019learning} and Occupancy Network~\cite{mescheder2019occupancy} learn the implicit fields to estimate the point-wise occupancy probability with a binary classifier.
To improve the effectiveness and generalization on complex scenes, latest studies propose to enhance implicit field by introducing extra inputs~\cite{xu2019disn,peng2020convolutional}, adopting advanced learning techniques~\cite{tretschk2020patchnets,duan2020curriculum,niemeyer2020differentiable,sitzmann2020metasdf} and decomposing the scene into local regions~\cite{chabra2020deep,takikawa2021neural,jiang2020local,genova2020local}.
NeRF~\cite{mildenhall2020nerf} represents complex scene by learning a view-dependent implicit neural radiance field, achieving high-resolution realistic novel view synthesis.


\section{Method}
\label{sec:method}

\paragraph{Overview.}
RayMVSNet estimates the depth maps from multiple overlapping RGB images. Similar to~\cite{yao2018mvsnet}, at each time, it takes one reference image $I_1$ and $N-1$ source images $\{I_i\}^N_2$ as input, and infers the depth map of the reference image.
RayMVSNet starts from building a light-weight 3D cost volume and estimating a coarse depth map (Sec.~\ref{sec:coarse}).
Then, epipolar transformer is proposed to learn the matching correlation of the pixel-wise 2D features of each view using attention mechanism (Sec.~\ref{sec:feature}).
The transformed features are fed into the 1D implicit field, implemented by an LSTM, along each camera viewing ray to estimate the signed distance functions (SDFs) of the hypothesized points as well as the zero-crossing position (Sec.~\ref{sec:ray}).
The method overview is illustrated in Figure \ref{fig:overview}.


\subsection{3D Cost Volume and Coarse Depth Prediction}
\label{sec:coarse}
We first feed the multi-view images $\{I_i\}^N_1$ to a 2D U-Net to extract image features $\{\bF^I_i\}^N_1$. The width and height of the image features are the same to those of the input images. Hence, $\{\bF^I_i\}^N_1$ preserve the fine appearance feature of local details, facilitating the high-resolution depth estimation.
By leveraging the 2D multi-view image features and the camera parameters, we build a variance-based 3D cost volume $V$, and extract the 3D volumetric features $\bF^{V}$ via a 3D U-Net~\cite{yao2018mvsnet}.
Since 3D convolution is memory-consuming, the resolution of $V$ in our work is set to be smaller than that in the previous works~\cite{cheng2020deep,yang2020cost,gu2020cascade}.
The coarse depth maps are estimated from the 3D volumetric features, which are then used for determining the modeling range of the ray-based 1D implicit fields.

\subsection{Epipolar Transformer}
\label{sec:feature}
We cast a set of rays $\mathbf{R}=\{\mathbf{r}_i\}^M_1$ from the camera's viewing direction of the reference image, where $M$ is the number of pixels in the reference image.
Our goal is to estimate the location of the zero-crossing point on each ray, so we can obtain the depth map of the reference view.
Compared to methods that estimate depth on the 3D cost volume, the ray-based method maintains the following advantages.
\emph{First}, since the depth map is view-dependent, ray-based depth optimization is more straightforward and light-weight.
\emph{Second}, all the ray-based 1D implicit fields share an identical spatial property, i.e. the monotonicity of the SDFs along the ray direction. As a result, the learning would be simplified and well regularized, leading to efficient network training and more accurate results.

\begin{figure}[t] \centering
	\begin{overpic}[width=0.97\linewidth,tics=10]{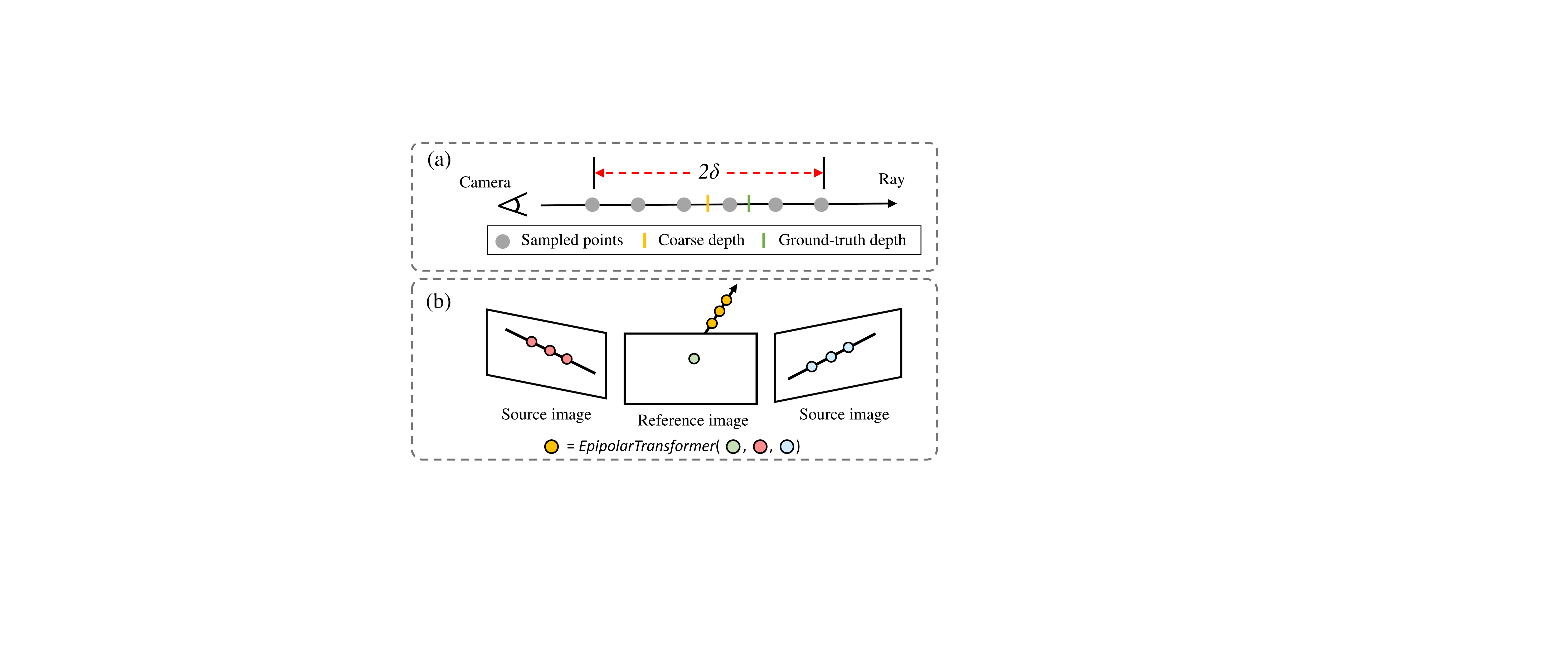}
   \end{overpic}\vspace{-5pt}
   \caption{
   (a) The hypothesized points are sampled around the predicted coarse depth to narrow down the search space of the zero-crossing position.
   (b) Epipolar transformer learns the matching correlation of the pixel-wise 2D features and aggregates these features using an attention mechanism.
   }
   \label{fig:method_details}\vspace{-5pt}
\end{figure}

\paragraph{Zero-crossing hypothesis sampling.}
We perform a point sampling to generate the zero-crossing point hypothesis on each ray.
Ideally, one could generate as many points as possible on each ray. However, most of the points are far from the surface, providing less informative information for the depth estimation.
To facilitate efficient training, as shown in Figure~\ref{fig:method_details} (a), we adopt the coarse depth map predicted in sec. \ref{sec:coarse} and uniformly sample $K$ points $P=\{p_k\}^K_1$ on the ray in the range of $\pm \delta$ around the estimated coarse depth.


\begin{figure}[t] \centering
	\begin{overpic}[width=0.97\linewidth,tics=10]{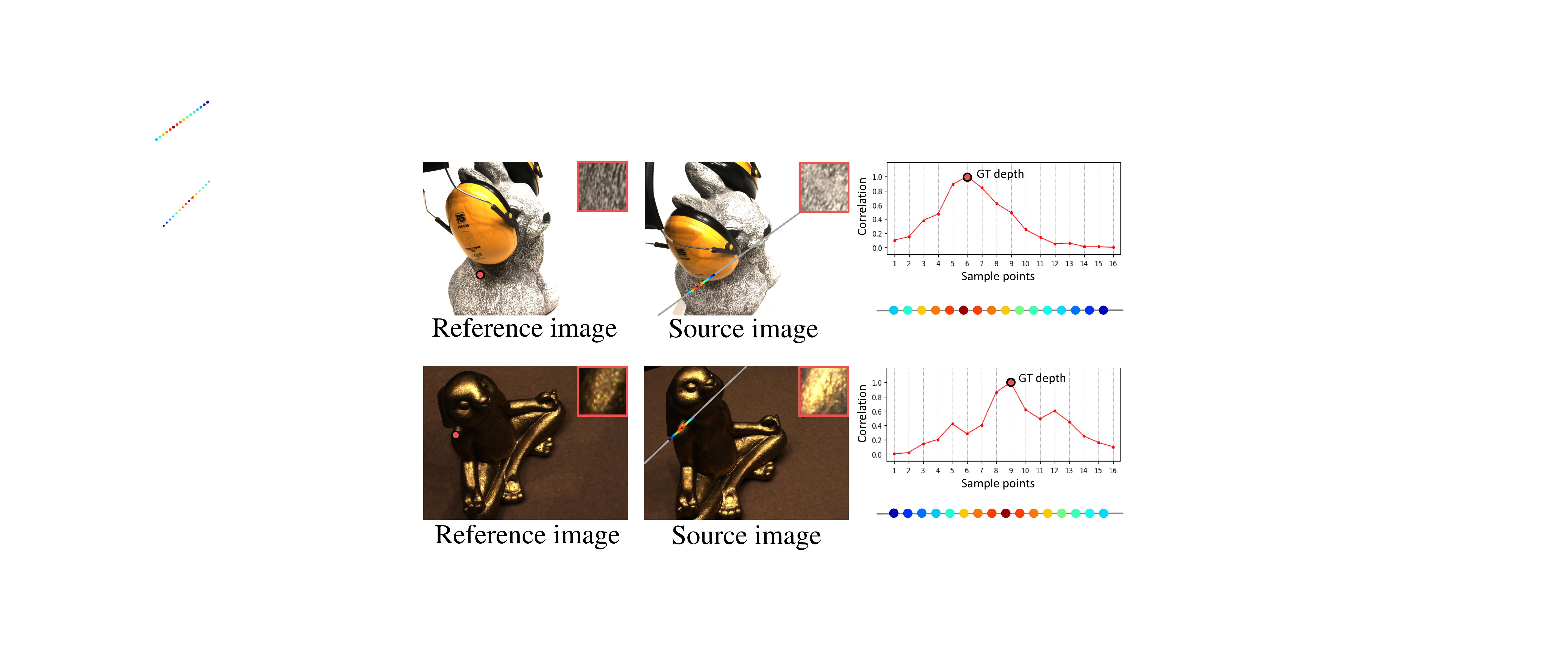}
   \end{overpic}\vspace{-5pt}
   \caption{
   Effects of epipolar transformer. Given a point in the reference image, epipolar transformer automatically selects reliable matching feature on the epipolar line of the source image. Note that it finds the matching feature correctly despite the influences of light changing (top row) and specular reflection (bottom row). The visualized point-pair correlations are deduced from the $\text{Softmax}(\bQ\bK^T)$ in Formulation \ref{eq:self_attention}.
   }
   \label{fig:epipolar}\vspace{-10pt}
\end{figure} 

\paragraph{Attention-aware cross-view feature correlation.}
The next step is to aggregate feature for the hypothesized points based on the multi-view image features.
A naive way to achieve this is to fetch the features from multi-view images based on the view projection, and take the variance.
However, image feature could be easily influenced by image defects, such as specular reflection and light changing. Naive variance considers all image features equally, which might incur unreliable features and provide incorrect cross-view feature correlation.
To alleviate this problem, we propose \emph{Epipolar Transformer} to learn cross-view feature correlation with attention mechanism (Figure~\ref{fig:method_details} b).

To be specific, the network architecture of epipolar transformer contains four self-attention layers, each followed by two AddNorm layers and one feed-forward layer.
Suppose $\bX=\text{Concat}(\bF^I_{1,p},...,\bF^I_{N,p})$, where $\text{Concat}(\cdot)$ is the concatenation operation, $\{\bF^I_{i,p}\}^N_1$ are the fetched multi-view image features at 3D point $p$. The self-attention layer of epipolar transformer is:
\begin{equation}\label{eq:self_attention}\small
\begin{aligned}
\bS = \text{SelfAttention}(\bQ, \bK, \bV) = \text{Softmax}(\bQ\bK^T)\bV,
\end{aligned}
\end{equation}
where $\bQ=\bX\bW^\bQ$, $\bK=\bX\bW^\bK$, $\bV=\bX\bW^\bV$ are the query vector, the key vector and the value vector respectively. $\bW^\bQ$, $\bW^\bK$, $\bW^\bV$ are the learned weights.
Examples to demonstrate the effects of first self-attention layer in epipolar transformer are visualized in Figure~\ref{fig:epipolar}.
The AddNorm layer of epipolar transformer is:
\begin{equation}\label{eq:addnorm}\small
\begin{aligned}
\bZ = \text{AddNorm}(\bX) = \text{LayerNorm}(\bX+\bS),
\end{aligned}
\end{equation}
where $\text{LayerNorm}(\cdot)$ is the layer normalization operation.
\revise{The output of epipolar transformer is the attention-aware denoised multi-view feature $\bF^A_{p}=\{\bF^A_{1,p},...,\bF^A_{N,p}\}$.}

\revise{
To further improve the feature quality, we concatenate the attention-aware feature with the 3D volume feature $\bF^{V}_{p}$ fetched from the 3D cost volume processed in Sec.~\ref{sec:coarse}:
\begin{equation}\label{eq:point_feature2}\small\vspace{-3pt}
\bF_p=\text{Concat}(\bF^A_{\mu,p},\bF^A_{\sigma,p},\bF^A_{1,p},\bF^V_{p}).
\end{equation}
where $\bF^A_{\mu,p}$ and $\bF^A_{\sigma,p}$ are the mean and variation of the elements in $\bF^A_{p}$~\cite{yao2018mvsnet,im2019dpsnet}. $\bF^A_{1,p}$ is the attention-aware feature at 3D point $p$ in the reference image.
}

\subsection{Ray-based 1D Implicit Field}
\label{sec:ray}

\paragraph{LSTM vs. alternative.}
Given the features of the hypothesized points, the ray-based 1D implicit fields are learned with an LSTM~\cite{hochreiter1997long}.
Crucially, we leverage two attributes of LSTM. \emph{First}, the mechanism of sequential processing inherently facilitates the learning of the SDF monotonicity along the ray direction. \emph{Second}, the property of time invariance increases the network robustness by allowing the zero-crossing position to appear at any place (time-step) on the ray.
An alternative to performing sequential inference is to use transformer~\cite{vaswani2017attention}. However, we experimentally found that replacing LSTM with transformer would not make the performance improve (see Table~\ref{tab:ablation}).

\begin{figure}[t] \centering
	\begin{overpic}[width=0.97\linewidth,tics=10]{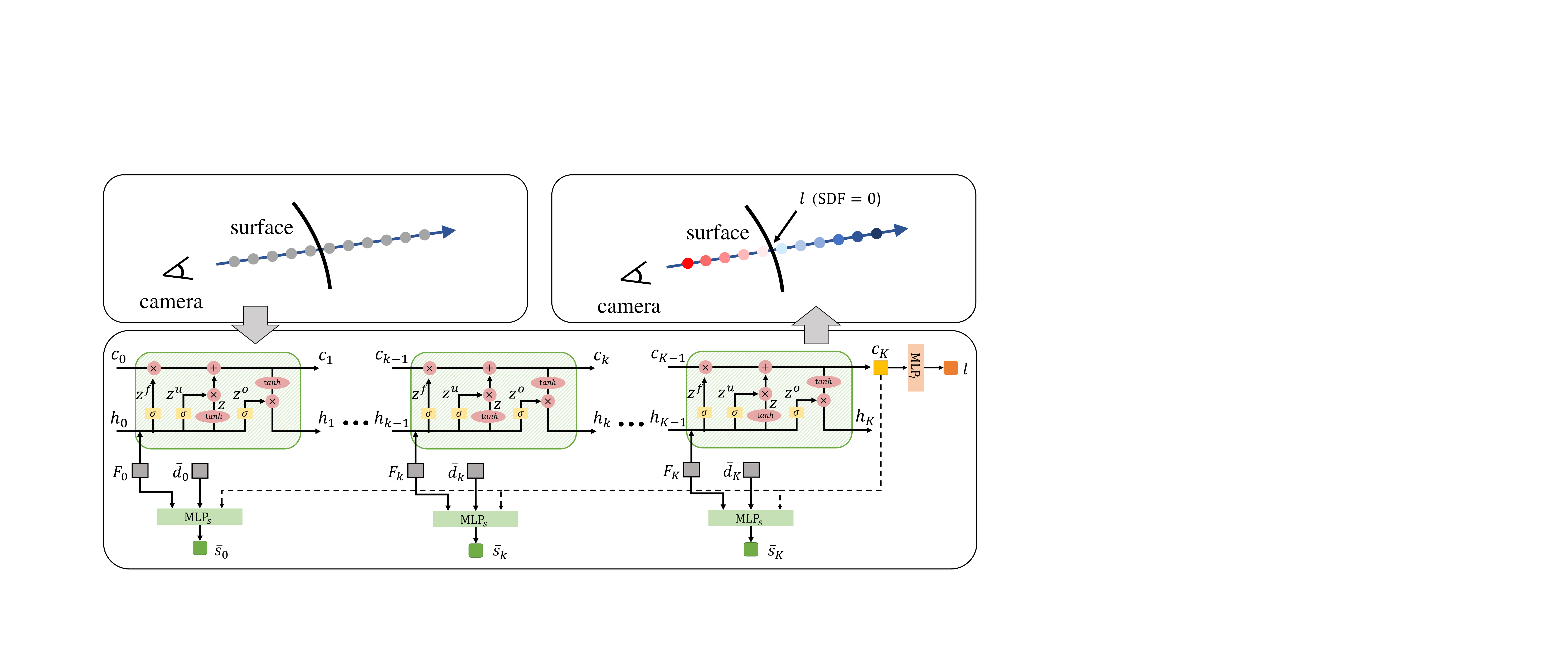}
   \end{overpic}\vspace{-5pt}
   \caption{Network architecture of the ray-based 1D implicit field. The hypothesized points are fed into an LSTM sequentially, to estimate the position of the zero-crossing point as well as the SDFs.}
   \label{fig:lstm}\vspace{-5pt}
\end{figure}

\paragraph{Network architecture.}
The network architecture of the 1D implicit field is shown in Figure~\ref{fig:lstm}.
The LSTM first aggregates the hypothesized points sequentially, and generates the ray feature $\bc_K$.
Specifically, the formulations of an LSTM unit at time-step $k$ are:
\begin{equation}\label{eq:lstm}\small\vspace{-2pt}
\begin{aligned}
&\bz=\text{tanh}(\bW[\bF_k,\bh_{k-1}]+b),\\
&\bz^f=\sigma(\bW^f[\bF_k,\bh_{k-1}]+b^f),\\
&\bz^u=\sigma(\bW^u[\bF_k,\bh_{k-1}]+b^u),\\
&\bz^o=\sigma(\bW^o[\bF_k,\bh_{k-1}]+b^o),\\
&\bc_k=\bz^f\circ \bc_{k-1}+\bz^u\circ \bz,\\
&\bh_k=\bz^o\circ \text{tanh}(\bc_k),\\
\end{aligned}
\end{equation}
where $\bF_k$ is the feature of point $p_k$, $\bh_k$ and $\bh_{k-1}$ are the hidden state of point $p_k$ and $p_{k-1}$ respectively, $\bz$ is the cell input activation vector, $\bz^f$ is the activation vector of the forget gate, $\bz^u$ is the activation vector of the update gate, $\bz^o$ is the activation vector of the output gate, $\bc_k$ is the cell state vector, $\bW$, $\bW^f$, $\bW^u$, $\bW^o$ are the weight matrices, $b$, $b^f$, $b^u$, $b^o$ are the weight vectors, $\circ$ is the element-wise multiplication. The LSTM is initialized with $\bc_0=0$ and $\bh_0=0$.

For each hypothesized point $p_k$, we use the ray feature $\bc_K$, the point-wise feature $\bF_k$ and its depth value $d_k$ (indicating the location on the ray) to estimate its SDF $s_k$ using an MLP.
Instead of using the true depth value $d_k$ and estimating the true SDF $s_k$, we use the normalized depth value $\overline{d}_k=k/K\in[0,1]$ and the normalized SDF $\overline{s}_k=s_k/s_{max}\in[-1,1]$, where $s_{max}$ is the maximal absolute SDF value on the ray.
Such normalization leads to a significant reduction of learning complexity and improvement of the result quality.
The formulation of the SDF prediction is:

\begin{equation}\label{eq:sdf}\small\vspace{-3pt}
\overline{s}_k=\text{MLP}_s([\bc_K, \bF_k, \overline{d}_k]).
\end{equation}

The above network predicts the SDFs of the hypothesized points on the ray. However, post-processing, e.g. ray casting, is still needed to find the zero-cross position. 
We extend our method to estimate the zero-cross position explicitly with another MLP.
Taking the ray feature $\bc_K$ as input, the MLP predicts the zero-crossing location $l\in[0,1]$ on the ray in the normalized 1D coordinate:

\begin{equation}\label{eq:ratio}\small\vspace{-3pt}
l=\text{MLP}_l(\bc_K).
\end{equation}



\paragraph{Loss functions.}
We adopt a multi-task learning strategy to optimize RayMVSNet. The two tasks, i.e. SDF estimation and zero-crossing position estimation, are inherently relevant and could reinforce each other by optimizing the following loss:

\begin{equation}\label{eq:loss_all}\small\vspace{-3pt}
\begin{aligned}
\mathcal{L} &= w_s\mathcal{L}_s + w_l\mathcal{L}_l + w_{sl}\mathcal{L}_{sl},\\
\end{aligned}
\end{equation}
where $\mathcal{L}_s$ and $\mathcal{L}_l$ are the loss of the SDF estimation and the zero-crossing location estimation, respectively:

\begin{equation}\label{eq:loss_s_l}\small\vspace{-2pt}
\begin{aligned}
\mathcal{L}_s &= \sum_{k=1}^K L_1(s_k,\hat{s_k}),\\
\mathcal{L}_l &= L_1(l,\hat{l}),\\
\end{aligned}
\end{equation}
where $\hat{s_k}$ and $\hat{l}$ are the ground-truth, $L_1(\cdot)$ denotes the L1 loss function. $\mathcal{L}_{sl}$ is a relational loss that penalizes the inconsistency between the predicted SDFs and the predicted zero-crossing position:
\begin{equation}\label{eq:loss_sl}\small\vspace{-2pt}
\begin{aligned}
\mathcal{L}_{sl} &=
\begin{cases}
1, & s_l^a \times s_l^b \textgreater 0\\
0, &  s_l^a \times s_l^b \leq 0,
\end{cases}
\end{aligned}
\end{equation}
where $s_l^a$ and $s_l^b$ are the predicted SDF of the closest two sampled points around the predicted zero-crossing position on the ray.
$w_s$, $w_l$, $w_{sl}$ are the pre-defined weights.



\begin{figure}[t] \centering
	\begin{overpic}[width=0.97\linewidth,tics=10]{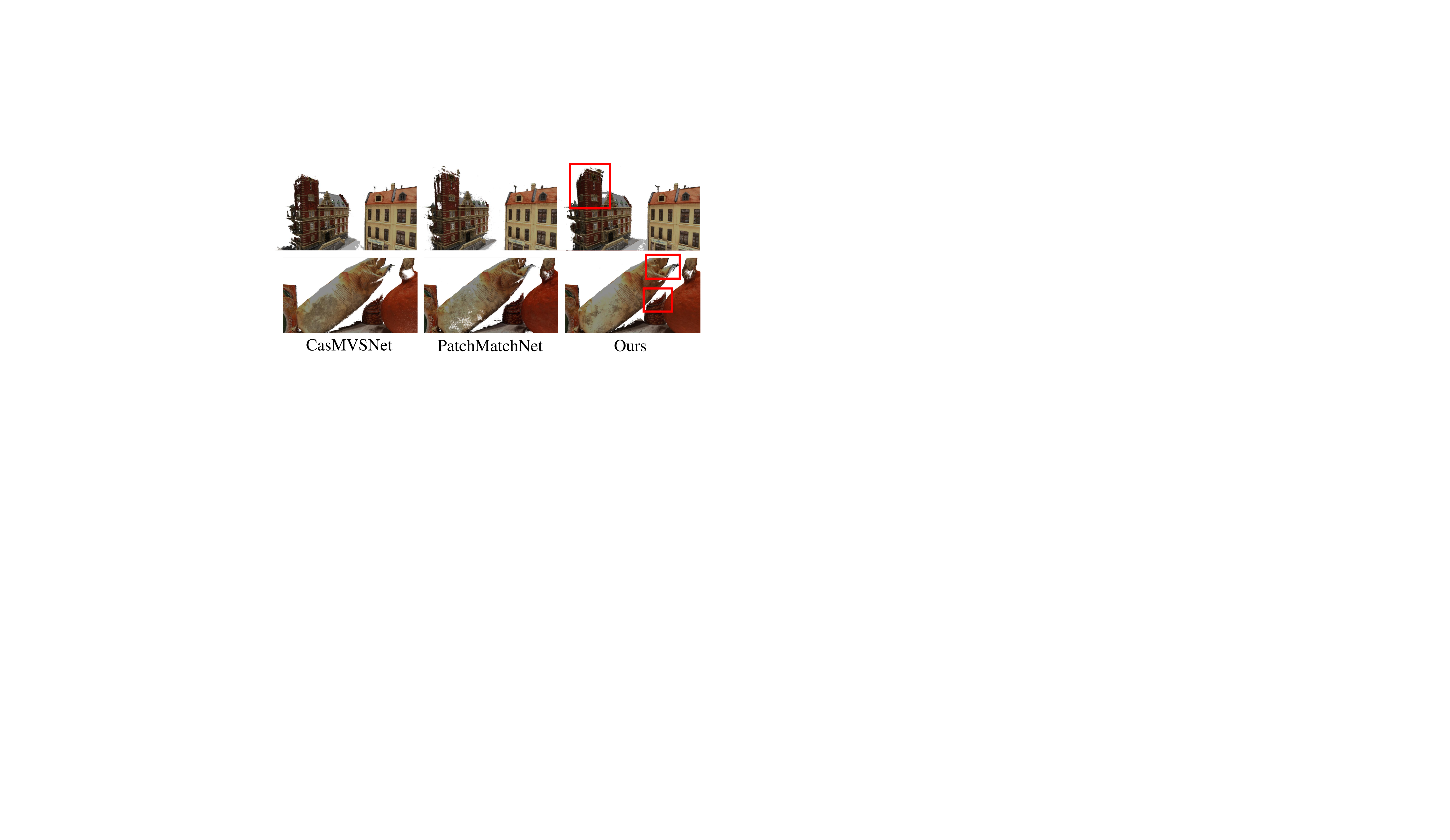}
   \end{overpic}\vspace{-5pt}
   \caption{Visual comparison of the reconstructed point cloud by RayMVSNet and the baselines.
   Please pay attention to the results of the challenging areas highlighted in the figures.}
   \label{fig:visual_comp}\vspace{-15pt}
\end{figure} 

\subsection{Implementations}
We provide implementation details of the training and inference.
At each time, RayMVSNet takes \ys{several images} with input image size $640\times 512$. The output feature size is $640\times 512\times 8$ .
The 2D U-Net consists of $6$ convolutional layers and $6$ deconvolutional layers, each followed by a batch normalization layer and a ReLU layer, except for the last ones.
The 3D cost volume is fed into a 3D U-Net which consists of three 3D convolutional layers and three 3D deconvolutional layers.
On each ray, the number of hypothesized points $K$ is $16$. Range of point sampling $\delta$ is $20mm$ for DTU and $100mm$ for Tanks \& Temples.
The feature fetching from images and volume are achieved by using bilinear interpolation and trilinear interpolation, respectively.
The hidden dimension of $\bz, \bz^f, \bz^u, \bz^o, \bc_k, \bh_k$ are $50$.
$\text{MLP}_l$ and $\text{MLP}_s$ both contain $4$ fully-convolutional layers. The weights $w_s$, $w_l$, $w_{sl}$ of multi-task learning loss function are $0.1$, $0.8$, $0.1$, respectively. Epipolar transformer and the LSTM are jointly trained. We use Adam optimizer with initial learning rate $0.0005$ which is decreased by $0.9$ for every $2$ epochs.
The training takes $48$ hours. The inference time is about \revise{$2$ seconds}.
We filter and fuse the depth maps to produce 3D point cloud like previous work~\cite{yao2018mvsnet}.

\section{Results and Evaluation}
\label{sec:result}

\paragraph{Datasets.}
We train and test RayMVSNet on the \emph{DTU} dataset~\cite{aanaes2016large}.
The \emph{DTU} dataset contains $79$ training scans and $22$ testing scans, all captured under changing lighting conditions.
Since \emph{DTU} did not provide SDF annotations, we densely generate the point-wise SDFs using the reconstructed surfaces~\cite{yao2018mvsnet,park2019deepsdf}.
Besides, three challenging test subsets focusing on regions with \emph{Specular reflection}, \emph{Shadow} and \emph{Occlusion} are created from the \emph{DTU} test set. These regions are manually annotated and are designed for evaluating the method performance on challenging cases. Please refer to the supplemental material for the subsets details.
To evaluate the generality, we test RayMVSNet on \emph{Tanks \& Temples}~\cite{knapitsch2017tanks} which contains large-scale complex scenes, using the trained model on \emph{DTU} without any fine-tuning.
\emph{BlendedMVS}~\cite{yao2020blendedmvs} is another large-scale dataset consists of various complex scenes. We provide qualitative results on \emph{BlendedMVS} to show the scalability of our method.

\subsection{Performance on DTU}

\paragraph{Evaluation on point cloud.}
To evaluate the proposed method on \emph{DTU},
we compare \emph{Accuracy} and \emph{Completeness} of the reconstructed point cloud using the distance metric in~\cite{aanaes2016large}.
The quantitative results are shown in Table \ref{tab:dtu}. It shows that our method not only produces competitive results in terms of \emph{Accuracy} and \emph{Completeness}, but also achieves the state-of-the-art \emph{Overall} performance.
This demonstrates the effectiveness of RayMVSNet, especially on balancing the trade-off between \emph{Accuracy} and \emph{Completeness}.
\ys{The qualitative comparisons are visualized in Figure \ref{fig:visual_comp}. It is shown that our method achieves high-quality reconstruction in various scenarios. In particular, our method outperforms the baselines in scenes with textureless regions, heavy occlusion, and complex geometry.}

\begin{table}[b]
\footnotesize\centering
\caption{
Quantitative results on the \emph{DTU} dataset. We compare all methods using the distance metric~\cite{aanaes2016large}. The numbers are reported in $mm$ (lower is better).
}\vspace{-10pt}
\label{tab:dtu}
\begin{tabular}{cccc}
\toprule
Method & Accuracy & Completeness & Overall \\ \midrule
Gipuma~\cite{galliani2015massively} & 0.283 & 0.873 & 0.578\\
MVSNet~\cite{yao2018mvsnet} & 0.396 & 0.527 & 0.462\\
R-MVSNet~\cite{yao2019recurrent} & 0.383 & 0.452 & 0.417\\
CIDER~\cite{xu2020learning} & 0.417 & 0.437 & 0.427\\
P-MVSNet~\cite{luo2019p} & 0.406 & 0.434 & 0.420\\
Point-MVSNet~\cite{chen2019point} & 0.342 & 0.411 & 0.376\\
Fast-MVSNet~\cite{yu2020fast} & 0.336 & 0.403 & 0.370\\
Att-MVSNet~\cite{luo2020attention} & 0.383 & 0.329 & 0.356\\
CasMVSNet~\cite{gu2020cascade} & 0.325 & 0.385 & 0.355\\
CVP-MVSNet~\cite{yang2020cost} & 0.296 & 0.406 & 0.351\\
PatchmatchNet~\cite{wang2021patchmatchnet} & 0.427 & 0.277 & 0.352\\
UCS-Net~\cite{cheng2020deep} & 0.338 & 0.349 & 0.344\\
AACVP-MVSNet~\cite{yu2021attention} & 0.357 & 0.326 & 0.341\\
U-MVS~\cite{xu2021digging} & 0.354 & 0.353 & 0.354\\ \midrule
Ours & 0.341 (6th)& 0.319 (2nd)& \textbf{0.330 (1st)}\\ \bottomrule
\end{tabular}
\vspace{1em}\vspace{-15pt}
\end{table}



\begin{figure}[t] \centering
	\begin{overpic}[width=0.97\linewidth,tics=10]{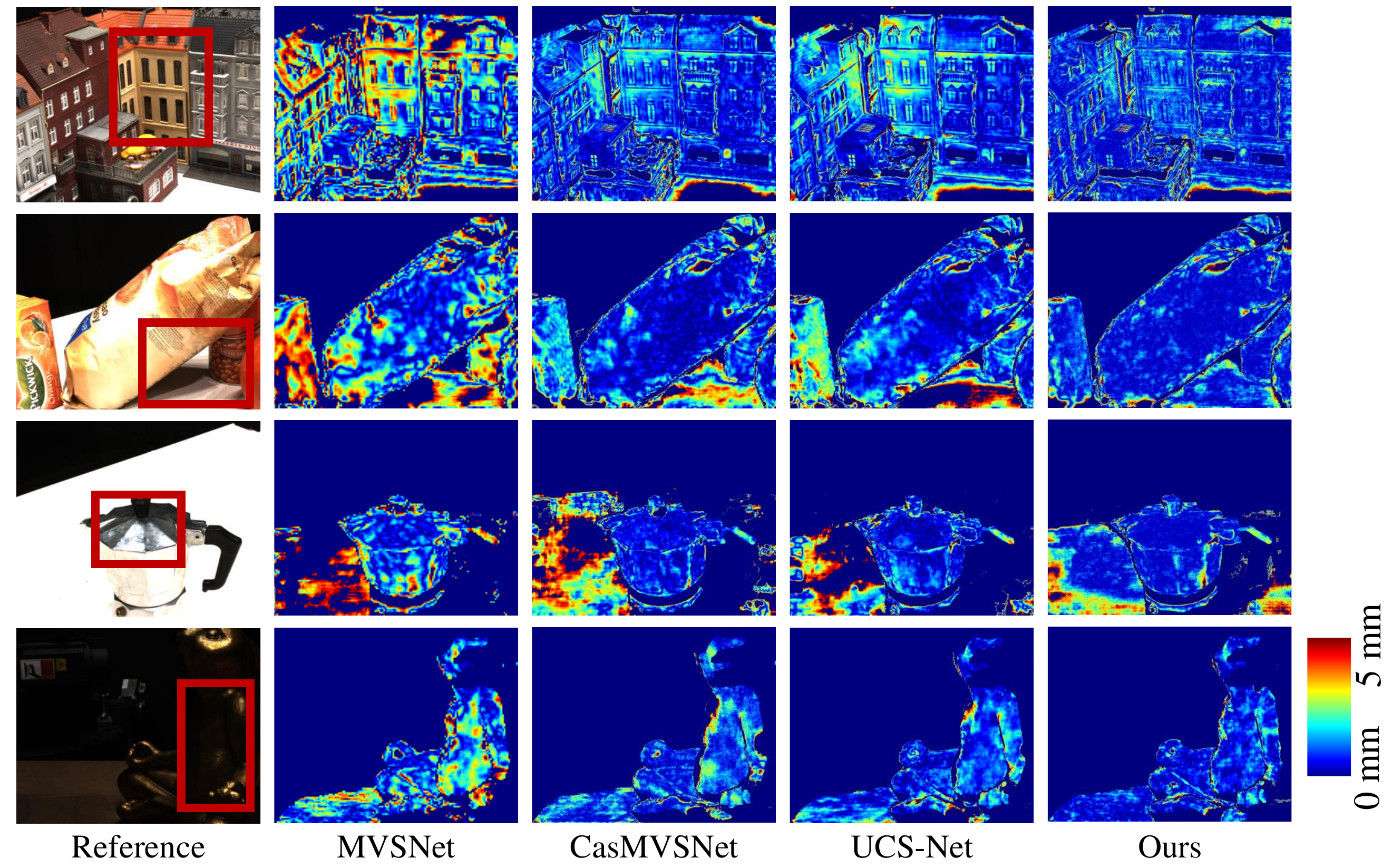}
   \end{overpic}\vspace{-5pt}
   \caption{
   \revise{Visual comparison of the estimated depth map by RayMVSNet and the baselines.}}
   \label{fig:depth}\vspace{-5pt}
\end{figure}

\begin{figure}[t] \centering
	\begin{overpic}[width=0.97\linewidth,tics=10]{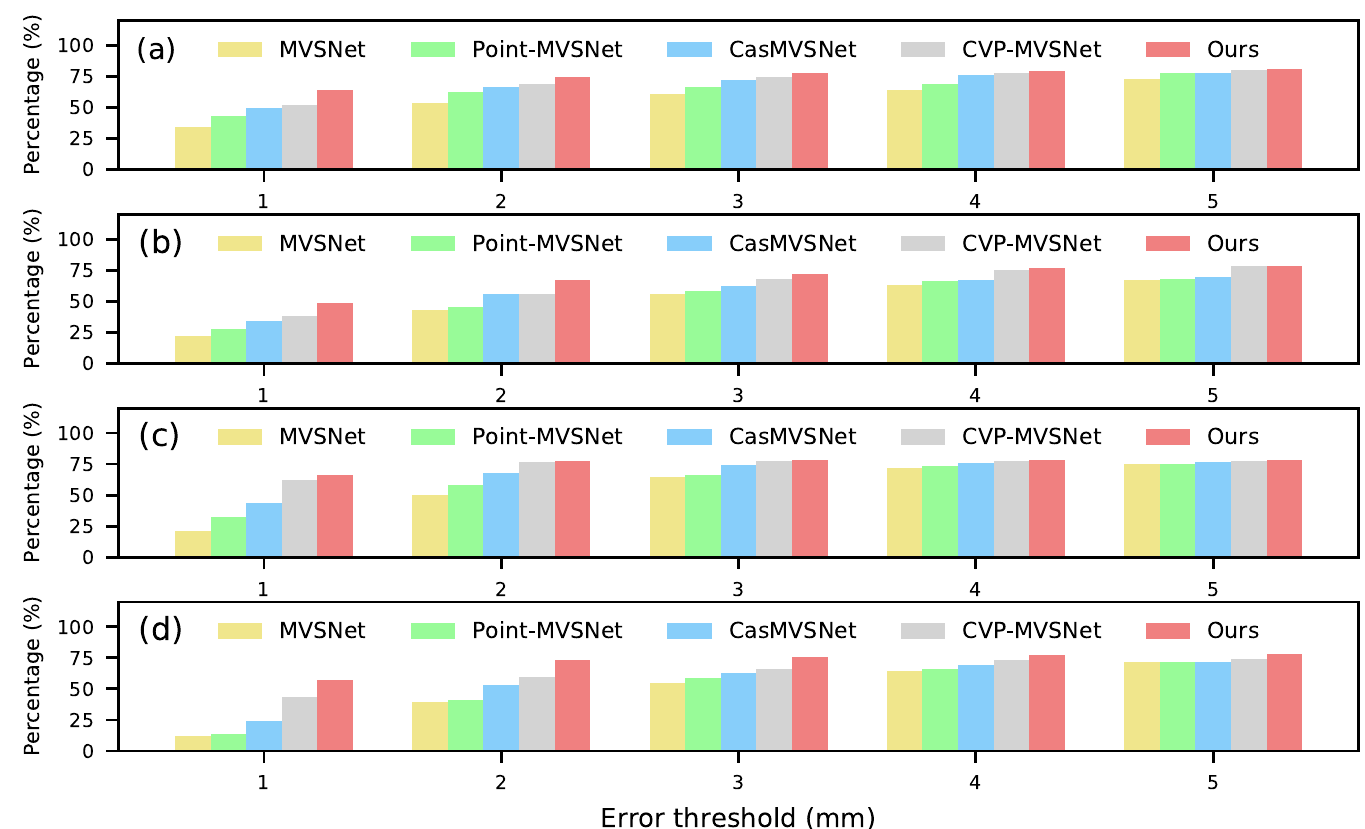}
   \end{overpic}\vspace{-5pt}
   \caption{Quantitative comparisons on the depth map prediction of the whole \emph{DTU} test set (a) and the challenging test subsets: \emph{Specular reflection} (b),  \emph{Shadow} (c) and \emph{Occlusion} (d).
   The percentage (Y-axis) represents the ratio of the pixels whose depth prediction error is smaller than the specific error thresholds (X-axis).}
   \label{fig:depth_curve}\vspace{-5pt}
\end{figure}

\paragraph{Evaluation on depth map.}
To further demonstrate our advantage, we compare RayMVSNet with existing works, in terms of the predicted depth map.
The quantitative comparisons on the whole \emph{DTU} test set (Figure \ref{fig:depth_curve} a) and the challenging subsets (Figure \ref{fig:depth_curve} b-d) are reported.
The percentage (Y-axis) represents the ratio of the pixels whose depth prediction error is smaller than the specific error thresholds (X-axis). Higher percentages represent better performances.
It is clear that our method outperforms all the baselines in all error thresholds.
\ys{Crucially, our method is more general and robust in challenging cases as shown in Figure \ref{fig:depth}, thanks to the prior learnt from the ray-based 1D implicit field.}

\begin{table*}
\footnotesize\centering
\caption{
Quantitative results on the \emph{Tanks \& temples} dataset. We use the f-score as the evaluation metric (higher is better).
}
\vspace{-10pt}
\label{tab:tat}
\begin{tabular}{p{3cm}<{\centering}p{1.05cm}<{\centering}p{1.05cm}<{\centering}p{1.05cm}<{\centering}p{1.5cm}<{\centering}p{1.05cm}<{\centering}p{1.05cm}<{\centering}p{1.05cm}<{\centering}p{1.05cm}<{\centering}p{1.05cm}<{\centering}}
\toprule
Method & Family & Francis & Horse & Light house & M60 & Panther & Playground & Train & Mean \\ \midrule
MVSNet~\cite{yao2018mvsnet} & 55.99 & 28.55 & 25.07 & 50.79 & 53.96 & 50.86 & 47.90 & 34.69 & 43.48 \\
R-MVSNet~\cite{yao2019recurrent} & 69.96 & 46.65 & 32.59 & 42.95 & 51.88 & 48.80 & 52.00 & 42.38 & 48.40 \\
PVA-MVSNet~\cite{yi2020pyramid} & 69.36 & 46.80 & 46.01 & 55.74 & 57.23 & 54.75 & 56.70 & 49.06 & 54.46 \\
CVP-MVSNet~\cite{yang2020cost} & 76.50 & 47.74 & 36.34 & 55.12 & 57.28 & 54.28 & 57.43 & 47.54 & 54.03 \\
CasMVSNet~\cite{gu2020cascade} & 76.37 & 58.45 & 46.26 & 55.81 & 56.11 & 54.06 & 58.18 & 49.51 & 56.84 \\
UCS-Net~\cite{cheng2020deep} & 76.09 & 53.16 &	43.03 & 54.00 & 55.60 & 51.49 & 57.38 & 47.89 & 54.83 \\
D2HC-RMVSNet~\cite{yan2020dense} & 74.69 & 56.04 & \textbf{49.42} & \textbf{60.08} & 59.81 & 59.61 & \textbf{60.04} & \textbf{53.92} & 59.20 \\
U-MVS~\cite{xu2021digging} & 76.49 & 60.04 & 49.20 & 55.52 & 55.33 & 51.22 & 56.77 & 52.63 & 57.15 \\ \midrule
Ours & \textbf{78.55} & \textbf{61.93} & 45.48 & 57.59 & \textbf{61.00} & \textbf{59.78} & 59.19 & 52.32 & \textbf{59.48}\\
\bottomrule
\end{tabular}
\end{table*}

\subsection{Performance on Tanks \& Temples}
We compare our method with the baselines on \emph{Tanks \& Temples}. Following the protocol of previous work~\cite{gu2020cascade}, we use the network trained on \emph{DTU}. \emph{F-score} is the evaluation metric. The quantitative results are shown in Table \ref{tab:tat}.
\ys{Our method achieves the best performance, demonstrating the generality of epipolar transformer and ray-based 1D implicit field on large-scale scenes.}

\subsection{Ablation Study}
In Table \ref{tab:ablation}, we conduct ablation studies to quantify the efficacy of several crucial components in RayMVSNet.

\begin{table}
\footnotesize\centering
\caption{
Ablation studies. The performance under distance metric is reported (lower is better).
}
\vspace{-10pt}
\label{tab:ablation}
\begin{tabular}{cccc}
\toprule
Method  & Accuracy & Completeness & Overall \\ \midrule
w/o epipolar transformer & 0.347 & 0.339 & 0.343 \\
w/o 2D image feature & 0.345 & 0.352 & 0.348 \\
w/o 3D volume feature & 0.434 & 0.322 & 0.378 \\
vis-max feature aggregation & 0.345 & 0.331 & 0.338 \\
Global implicit field & 0.573 & 0.642 & 0.608 \\
Ray with Transformer & \textbf{0.339} & 0.343 & 0.341  \\
Ray with average pooling & 0.356 & 0.406 & 0.381 \\
Ray with max pooling & 0.466 & 0.383 & 0.424 \\
w/o SDF prediction & 0.354 & 0.330 & 0.342 \\ \midrule
Ours & 0.341 & \textbf{0.319} & \textbf{0.330} \\
\bottomrule
\end{tabular}
\end{table}

\paragraph{Feature aggregation.}
The cross-view feature aggregation is a key component of RayMVSNet. To evaluate the importance, we compare the full method to several baselines without some specific component: \emph{w/o epipolar transformer}, \emph{w/o 2D image feature} and \emph{w/o 3D volume feature}.
\ys{It clearly shows that all these baselines make the performance decline.
\emph{It is worth noting that w/o epipolar transformer} achieves a lower completeness score, indicating epipolar transformer could make the reconstruction complete by providing more reliable cross-view correlations.}
\revise{We also compare our epipolar transformer to other multi-view feature aggregation method. In the experiment of \emph{vis-max feature aggregation}, we replace the epipolar transformer with the visibility-aware max-pooling feature aggregation~\cite{chen2020visibility}. The result indicates epipolar transformer is a better solution.}


\begin{figure}[t] \centering
	\begin{overpic}[width=0.97\linewidth,tics=10]{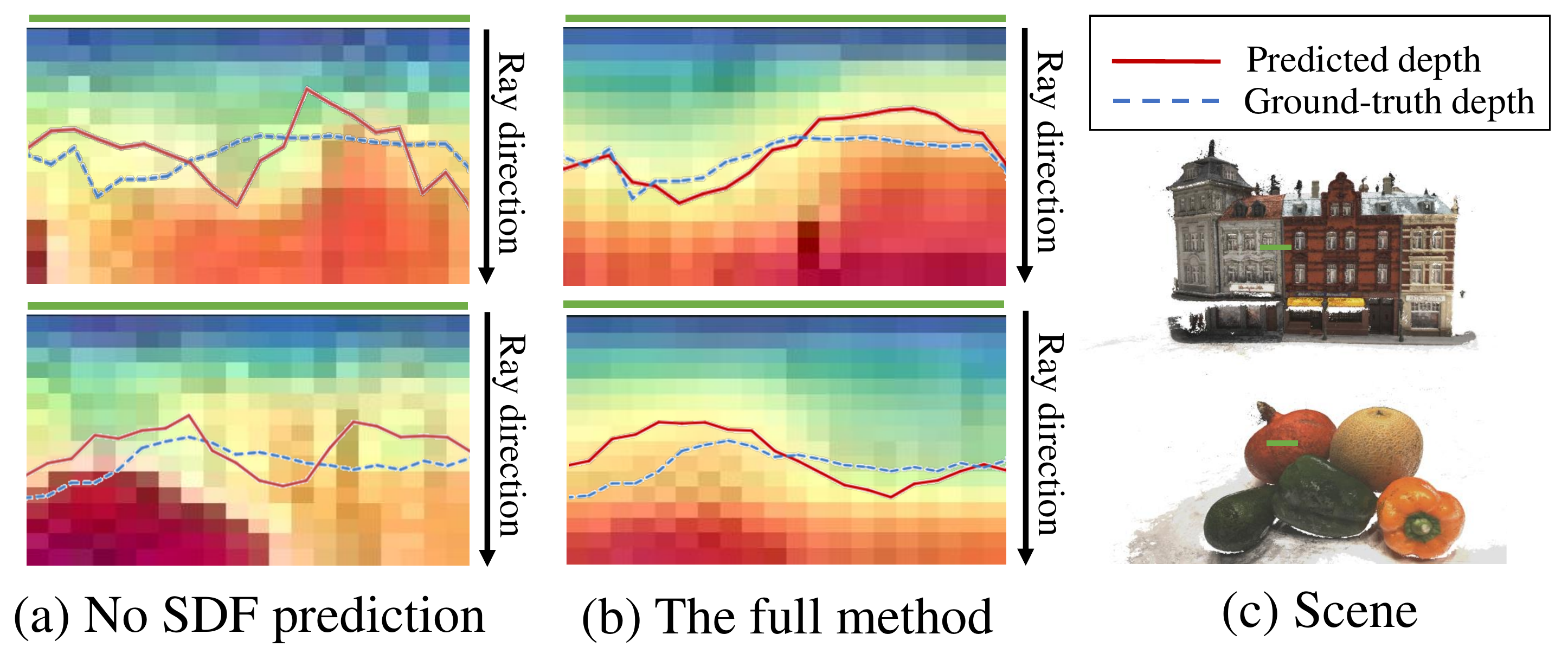}
   \end{overpic}\vspace{-5pt}
   \caption{Mid-layer feature map t-SNE visualization of the w/o SDF prediction baseline (a) and the full method (b) for the green segment marked in the scenes in (c). }
   \label{fig:sdf}\vspace{-5pt}
\end{figure} 


\begin{figure*}[!t] \centering\vspace{-7pt} \begin{overpic}[width=0.97\linewidth,tics=10]{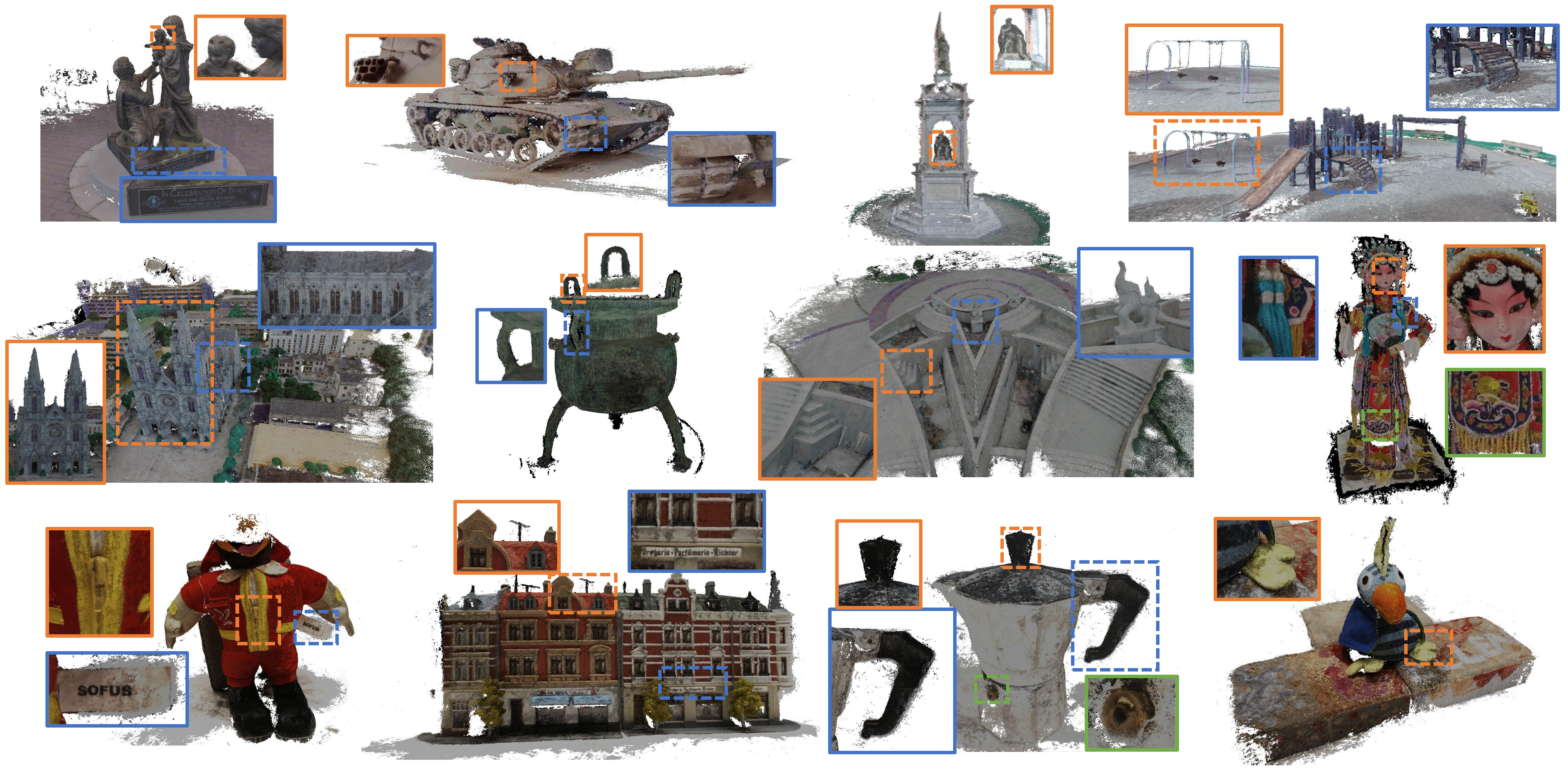}
   \end{overpic}\vspace{-5pt}
   \caption{Gallery of the reconstructed point cloud on \emph{Tanks \& temples} (top row), \emph{BlendedMVS} (middle row) and \emph{DTU} (bottom row).}
   \label{fig:gallery}\vspace{-7pt}
\end{figure*} 


\begin{figure}[t] \centering
	\begin{overpic}[width=0.97\linewidth,tics=10]{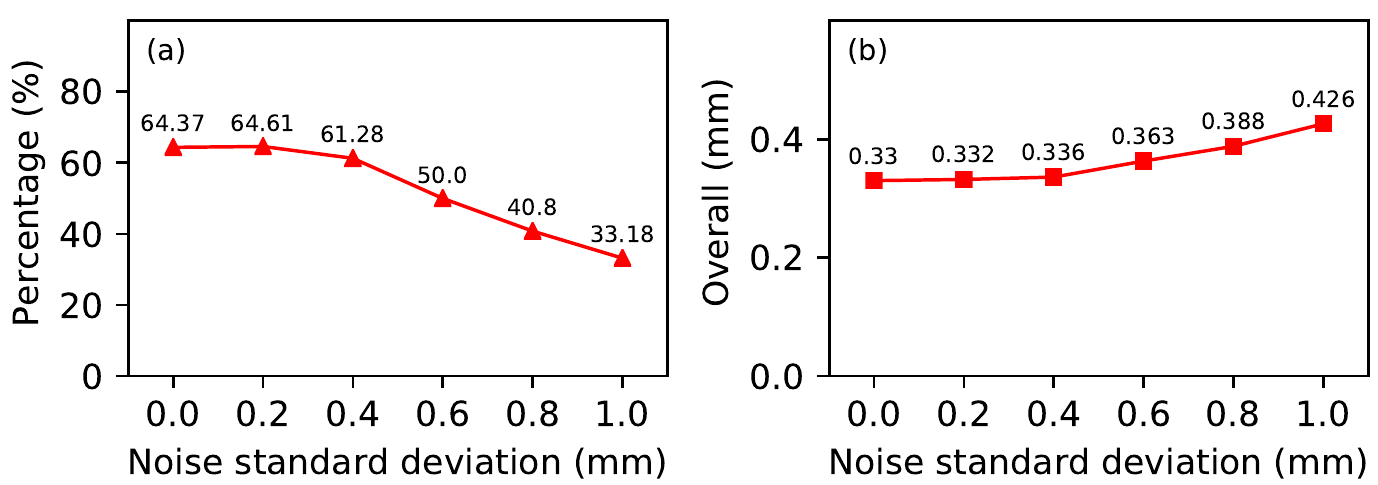}
   \end{overpic}\vspace{-5pt}
   \caption{Sensitivity to coarse depth quality. The percentage of pixel-wise depth predictions whose error is smaller than $1mm$ (a) and the overall score of point cloud reconstruction (b) are reported. }
   \label{fig:noise}\vspace{-10pt}
\end{figure} 

\paragraph{Global implicit field.}
Our method learns the local 1D implicit field by splitting the scene into a bunch of rays. To show its necessity, a straightforward baseline is to learn a global implicit field in the reference frustum directly, such that there is no ray-based representation. This baseline adopts the same cross-view feature aggregation as the full method, and predicts the SDF of all points in the reference frustum by using an MLP. The depth map is then generated by a ray-casting algorithm from the predicted SDFs. \ys{Unsurprisingly, experiments show this network is hard to converge and leads to low quantitative performance, which suggests that the ray-based 1D implicit field indeed simplifies the learning and is suitable to the MVS problem.}

\paragraph{Other ray-based implicit field models.}
In order to reveal the need of the proposed LSTM, we compare our method against several baselines with alternative models of processing sequential data. To be specific, we study the effects of replacing the LSTM with average pooling, max pooling, and Transformer~\cite{vaswani2017attention}, respectively. The \emph{Ray with average pooling} and the \emph{Ray with max pooling} baselines aggregate ray feature by average pooling and max pooling over all sampled points, respectively. The aggregated features are then used to predict the zero-crossing location. The point-wise SDF predictions are also performed as an auxiliary task.
The result shows that our method outperforms all the baselines.
\ys{In particular, the performance drops significantly with the \emph{Ray with average pooling} and the \emph{Ray with max pooling}, implying that the modeling of ray-based 1D implicit field is a non-trivial task.
The \emph{Ray with Transformer} is inferior to the full method, in terms of the \emph{Overall score}, confirming that LSTM is more appropriate to our problem.}


\paragraph{No SDF prediction.}
The SDF prediction is an auxiliary task in RayMVSNet. We demonstrate its influence by turning it off and comparing to the full method. \ys{The performance of \emph{w/o SDF prediction} baseline is inferior to the full method, demonstrating the joint training of SDF prediction and zero-crossing position prediction is indeed helpful, due to the extra supervision of SDF. Examples are visualized in Figure \ref{fig:sdf} which compares the mid-layer features of the full model and the baseline without SDF prediction. We can see that the mid-layer features of the full method, with SDF supervision, maintain a better monotonicity along the ray direction, resulting in more accurate predictions.}

\subsection{Sensitivity to coarse depth quality}
We show our RayMVSNet is robust to the incorrectness of coarse depth prediction with a pressure test. In the experiment, we add Gaussian noise to the predicted coarse depth maps, during both the training and testing phases. We report the performance of the depth map prediction and the point cloud reconstruction on \emph{DTU}. Figure~\ref{fig:noise} shows RayMVSNet is robust to moderate perturbation (noise standard deviation $\leq0.4mm$). It is interesting to see that the quality of depth map prediction slightly increases when moderate noise is added. This demonstrates that data augmentation such as modest perturbation to coarse depth is helpful for training a more generalizable RayMVSNet.

\subsection{Qualitative results}
We visualize the qualitative results of RayMVSNet on several datasets in Figure \ref{fig:gallery}. Note that RayMVSNet is able to reconstruct large-scale scenes with fine-grained geometry details, such as the highlighted regions.


\section{Conclusion}
\label{sec:conclusion}

We have presented RayMVSNet, which learns to directly optimize the depth value along each camera ray. An epipolar transformer is designed to enable sequential modeling of 1D ray-based implicit fields, which essentially mimics the epipolar line search in traditional MVS. The ray-based approach demonstrates significant performance boost with only a low-res cost volume. An interesting future direction is to further enhance the ray-based deep MVS approach so that cost volume convolution could be completely saved. In most deep MVS works, 3D point cloud is recovered from the estimated depth map as a post-processing. We would like to study end-to-end optimization of 3D point clouds~\cite{shi2020symmetrynet}.

\section*{Acknowledgements}
We thank the anonymous reviewers for their valuable comments. This work was supported in part by NSFC (62132021, 62102435, 62002379, U20A20185, 61972435), National Key Research and Development Program of China (2018AAA0102200) and the Zhejiang Lab’s International Talent Fund for Young Professionals.

{\small
\bibliographystyle{ieee_fullname}
\bibliography{egbib}

\begin{thebibliography}{10}\itemsep=-1pt

\bibitem{aanaes2016large}
Henrik Aan{\ae}s, Rasmus~Ramsb{\o}l Jensen, George Vogiatzis, Engin Tola, and
  Anders~Bjorholm Dahl.
\newblock Large-scale data for multiple-view stereopsis.
\newblock {\em International Journal of Computer Vision}, 120(2):153--168,
  2016.

\bibitem{andrew2001multiple}
Alex~M Andrew.
\newblock Multiple view geometry in computer vision.
\newblock {\em Kybernetes}, 2001.

\bibitem{chabra2020deep}
Rohan Chabra, Jan~E Lenssen, Eddy Ilg, Tanner Schmidt, Julian Straub, Steven
  Lovegrove, and Richard Newcombe.
\newblock Deep local shapes: Learning local sdf priors for detailed 3d
  reconstruction.
\newblock In {\em European Conference on Computer Vision}, pages 608--625.
  Springer, 2020.

\bibitem{chen2021mvsnerf}
Anpei Chen, Zexiang Xu, Fuqiang Zhao, Xiaoshuai Zhang, Fanbo Xiang, Jingyi Yu,
  and Hao Su.
\newblock Mvsnerf: Fast generalizable radiance field reconstruction from
  multi-view stereo.
\newblock {\em arXiv preprint arXiv:2103.15595}, 2021.

\bibitem{chen2019point}
Rui Chen, Songfang Han, Jing Xu, and Hao Su.
\newblock Point-based multi-view stereo network.
\newblock In {\em Proceedings of the IEEE/CVF International Conference on
  Computer Vision}, pages 1538--1547, 2019.

\bibitem{chen2020visibility}
Rui Chen, Songfang Han, Jing Xu, and Hao Su.
\newblock Visibility-aware point-based multi-view stereo network.
\newblock {\em IEEE transactions on pattern analysis and machine intelligence},
  43(10):3695--3708, 2020.

\bibitem{chen2019learning}
Zhiqin Chen and Hao Zhang.
\newblock Learning implicit fields for generative shape modeling.
\newblock In {\em Proceedings of the IEEE/CVF Conference on Computer Vision and
  Pattern Recognition}, pages 5939--5948, 2019.

\bibitem{cheng2020deep}
Shuo Cheng, Zexiang Xu, Shilin Zhu, Zhuwen Li, Li~Erran Li, Ravi Ramamoorthi,
  and Hao Su.
\newblock Deep stereo using adaptive thin volume representation with
  uncertainty awareness.
\newblock In {\em Proceedings of the IEEE/CVF Conference on Computer Vision and
  Pattern Recognition}, pages 2524--2534, 2020.

\bibitem{deng2021deformed}
Yu Deng, Jiaolong Yang, and Xin Tong.
\newblock Deformed implicit field: Modeling 3d shapes with learned dense
  correspondence.
\newblock In {\em Proceedings of the IEEE/CVF Conference on Computer Vision and
  Pattern Recognition}, pages 10286--10296, 2021.

\bibitem{duan2020curriculum}
Yueqi Duan, Haidong Zhu, He Wang, Li Yi, Ram Nevatia, and Leonidas~J Guibas.
\newblock Curriculum deepsdf.
\newblock In {\em European Conference on Computer Vision}, pages 51--67.
  Springer, 2020.

\bibitem{galliani2015massively}
Silvano Galliani, Katrin Lasinger, and Konrad Schindler.
\newblock Massively parallel multiview stereopsis by surface normal diffusion.
\newblock In {\em Proceedings of the IEEE International Conference on Computer
  Vision}, pages 873--881, 2015.

\bibitem{genova2020local}
Kyle Genova, Forrester Cole, Avneesh Sud, Aaron Sarna, and Thomas Funkhouser.
\newblock Local deep implicit functions for 3d shape.
\newblock In {\em Proceedings of the IEEE/CVF Conference on Computer Vision and
  Pattern Recognition}, pages 4857--4866, 2020.

\bibitem{gu2020cascade}
Xiaodong Gu, Zhiwen Fan, Siyu Zhu, Zuozhuo Dai, Feitong Tan, and Ping Tan.
\newblock Cascade cost volume for high-resolution multi-view stereo and stereo
  matching.
\newblock In {\em Proceedings of the IEEE/CVF Conference on Computer Vision and
  Pattern Recognition}, pages 2495--2504, 2020.

\bibitem{hartmann2017learned}
Wilfried Hartmann, Silvano Galliani, Michal Havlena, Luc Van~Gool, and Konrad
  Schindler.
\newblock Learned multi-patch similarity.
\newblock In {\em Proceedings of the IEEE International Conference on Computer
  Vision}, pages 1586--1594, 2017.

\bibitem{hochreiter1997long}
Sepp Hochreiter and J{\"u}rgen Schmidhuber.
\newblock Long short-term memory.
\newblock {\em Neural computation}, 9(8):1735--1780, 1997.

\bibitem{huang2018deepmvs}
Po-Han Huang, Kevin Matzen, Johannes Kopf, Narendra Ahuja, and Jia-Bin Huang.
\newblock Deepmvs: Learning multi-view stereopsis.
\newblock In {\em Proceedings of the IEEE Conference on Computer Vision and
  Pattern Recognition}, pages 2821--2830, 2018.

\bibitem{im2019dpsnet}
Sunghoon Im, Hae-Gon Jeon, Stephen Lin, and In~So Kweon.
\newblock Dpsnet: End-to-end deep plane sweep stereo.
\newblock {\em arXiv preprint arXiv:1905.00538}, 2019.

\bibitem{ji2017surfacenet}
Mengqi Ji, Juergen Gall, Haitian Zheng, Yebin Liu, and Lu Fang.
\newblock Surfacenet: An end-to-end 3d neural network for multiview stereopsis.
\newblock In {\em Proceedings of the IEEE International Conference on Computer
  Vision}, pages 2307--2315, 2017.

\bibitem{jiang2020local}
Chiyu Jiang, Avneesh Sud, Ameesh Makadia, Jingwei Huang, Matthias Nie{\ss}ner,
  Thomas Funkhouser, et~al.
\newblock Local implicit grid representations for 3d scenes.
\newblock In {\em Proceedings of the IEEE/CVF Conference on Computer Vision and
  Pattern Recognition}, pages 6001--6010, 2020.

\bibitem{knapitsch2017tanks}
Arno Knapitsch, Jaesik Park, Qian-Yi Zhou, and Vladlen Koltun.
\newblock Tanks and temples: Benchmarking large-scale scene reconstruction.
\newblock {\em ACM Transactions on Graphics (ToG)}, 36(4):1--13, 2017.

\bibitem{luo2019p}
Keyang Luo, Tao Guan, Lili Ju, Haipeng Huang, and Yawei Luo.
\newblock P-mvsnet: Learning patch-wise matching confidence aggregation for
  multi-view stereo.
\newblock In {\em Proceedings of the IEEE/CVF International Conference on
  Computer Vision}, pages 10452--10461, 2019.

\bibitem{luo2020attention}
Keyang Luo, Tao Guan, Lili Ju, Yuesong Wang, Zhuo Chen, and Yawei Luo.
\newblock Attention-aware multi-view stereo.
\newblock In {\em Proceedings of the IEEE/CVF Conference on Computer Vision and
  Pattern Recognition}, pages 1590--1599, 2020.

\bibitem{ma2021epp}
Xinjun Ma, Yue Gong, Qirui Wang, Jingwei Huang, Lei Chen, and Fan Yu.
\newblock Epp-mvsnet: Epipolar-assembling based depth prediction for multi-view
  stereo.
\newblock In {\em Proceedings of the IEEE/CVF International Conference on
  Computer Vision}, pages 5732--5740, 2021.

\bibitem{mescheder2019occupancy}
Lars Mescheder, Michael Oechsle, Michael Niemeyer, Sebastian Nowozin, and
  Andreas Geiger.
\newblock Occupancy networks: Learning 3d reconstruction in function space.
\newblock In {\em Proceedings of the IEEE/CVF Conference on Computer Vision and
  Pattern Recognition}, pages 4460--4470, 2019.

\bibitem{mildenhall2020nerf}
Ben Mildenhall, Pratul~P Srinivasan, Matthew Tancik, Jonathan~T Barron, Ravi
  Ramamoorthi, and Ren Ng.
\newblock Nerf: Representing scenes as neural radiance fields for view
  synthesis.
\newblock In {\em European conference on computer vision}, pages 405--421.
  Springer, 2020.

\bibitem{nelson2020learning}
Henry~J Nelson and Nikolaos Papanikolopoulos.
\newblock Learning continuous object representations from point cloud data.
\newblock In {\em 2020 IEEE/RSJ International Conference on Intelligent Robots
  and Systems (IROS)}, pages 2446--2451. IEEE, 2020.

\bibitem{niemeyer2020differentiable}
Michael Niemeyer, Lars Mescheder, Michael Oechsle, and Andreas Geiger.
\newblock Differentiable volumetric rendering: Learning implicit 3d
  representations without 3d supervision.
\newblock In {\em Proceedings of the IEEE/CVF Conference on Computer Vision and
  Pattern Recognition}, pages 3504--3515, 2020.

\bibitem{park2019deepsdf}
Jeong~Joon Park, Peter Florence, Julian Straub, Richard Newcombe, and Steven
  Lovegrove.
\newblock Deepsdf: Learning continuous signed distance functions for shape
  representation.
\newblock In {\em Proceedings of the IEEE/CVF Conference on Computer Vision and
  Pattern Recognition}, pages 165--174, 2019.

\bibitem{peng2020convolutional}
Songyou Peng, Michael Niemeyer, Lars Mescheder, Marc Pollefeys, and Andreas
  Geiger.
\newblock Convolutional occupancy networks.
\newblock In {\em Computer Vision--ECCV 2020: 16th European Conference,
  Glasgow, UK, August 23--28, 2020, Proceedings, Part III 16}, pages 523--540.
  Springer, 2020.

\bibitem{shi2020symmetrynet}
Yifei Shi, Junwen Huang, Hongjia Zhang, Xin Xu, Szymon Rusinkiewicz, and Kai
  Xu.
\newblock Symmetrynet: learning to predict reflectional and rotational
  symmetries of 3d shapes from single-view rgb-d images.
\newblock {\em ACM Transactions on Graphics (TOG)}, 39(6):1--14, 2020.

\bibitem{sitzmann2020metasdf}
Vincent Sitzmann, Eric~R Chan, Richard Tucker, Noah Snavely, and Gordon
  Wetzstein.
\newblock Metasdf: Meta-learning signed distance functions.
\newblock {\em arXiv preprint arXiv:2006.09662}, 2020.

\bibitem{takikawa2021neural}
Towaki Takikawa, Joey Litalien, Kangxue Yin, Karsten Kreis, Charles Loop, Derek
  Nowrouzezahrai, Alec Jacobson, Morgan McGuire, and Sanja Fidler.
\newblock Neural geometric level of detail: Real-time rendering with implicit
  3d shapes.
\newblock In {\em Proceedings of the IEEE/CVF Conference on Computer Vision and
  Pattern Recognition}, pages 11358--11367, 2021.

\bibitem{tretschk2020patchnets}
Edgar Tretschk, Ayush Tewari, Vladislav Golyanik, Michael Zollh{\"o}fer,
  Carsten Stoll, and Christian Theobalt.
\newblock Patchnets: Patch-based generalizable deep implicit 3d shape
  representations.
\newblock In {\em European Conference on Computer Vision}, pages 293--309.
  Springer, 2020.

\bibitem{vaswani2017attention}
Ashish Vaswani, Noam Shazeer, Niki Parmar, Jakob Uszkoreit, Llion Jones,
  Aidan~N Gomez, {\L}ukasz Kaiser, and Illia Polosukhin.
\newblock Attention is all you need.
\newblock In {\em Advances in neural information processing systems}, pages
  5998--6008, 2017.

\bibitem{wang2021patchmatchnet}
Fangjinhua Wang, Silvano Galliani, Christoph Vogel, Pablo Speciale, and Marc
  Pollefeys.
\newblock Patchmatchnet: Learned multi-view patchmatch stereo.
\newblock In {\em Proceedings of the IEEE/CVF Conference on Computer Vision and
  Pattern Recognition}, pages 14194--14203, 2021.

\bibitem{wei2021aa}
Zizhuang Wei, Qingtian Zhu, Chen Min, Yisong Chen, and Guoping Wang.
\newblock Aa-rmvsnet: Adaptive aggregation recurrent multi-view stereo network.
\newblock In {\em Proceedings of the IEEE/CVF International Conference on
  Computer Vision}, pages 6187--6196, 2021.

\bibitem{xu2021digging}
Hongbin Xu, Zhipeng Zhou, Yali Wang, Wenxiong Kang, Baigui Sun, Hao Li, and Yu
  Qiao.
\newblock Digging into uncertainty in self-supervised multi-view stereo.
\newblock In {\em Proceedings of the IEEE/CVF International Conference on
  Computer Vision}, pages 6078--6087, 2021.

\bibitem{xu2020learning}
Qingshan Xu and Wenbing Tao.
\newblock Learning inverse depth regression for multi-view stereo with
  correlation cost volume.
\newblock In {\em Proceedings of the AAAI Conference on Artificial
  Intelligence}, volume~34, pages 12508--12515, 2020.

\bibitem{xu2019disn}
Qiangeng Xu, Weiyue Wang, Duygu Ceylan, Radomir Mech, and Ulrich Neumann.
\newblock Disn: Deep implicit surface network for high-quality single-view 3d
  reconstruction.
\newblock {\em arXiv preprint arXiv:1905.10711}, 2019.

\bibitem{yan2020dense}
Jianfeng Yan, Zizhuang Wei, Hongwei Yi, Mingyu Ding, Runze Zhang, Yisong Chen,
  Guoping Wang, and Yu-Wing Tai.
\newblock Dense hybrid recurrent multi-view stereo net with dynamic consistency
  checking.
\newblock In {\em European Conference on Computer Vision}, pages 674--689.
  Springer, 2020.

\bibitem{yang2020cost}
Jiayu Yang, Wei Mao, Jose~M Alvarez, and Miaomiao Liu.
\newblock Cost volume pyramid based depth inference for multi-view stereo.
\newblock In {\em Proceedings of the IEEE/CVF Conference on Computer Vision and
  Pattern Recognition}, pages 4877--4886, 2020.

\bibitem{yao2018mvsnet}
Yao Yao, Zixin Luo, Shiwei Li, Tian Fang, and Long Quan.
\newblock Mvsnet: Depth inference for unstructured multi-view stereo.
\newblock In {\em Proceedings of the European Conference on Computer Vision
  (ECCV)}, pages 767--783, 2018.

\bibitem{yao2019recurrent}
Yao Yao, Zixin Luo, Shiwei Li, Tianwei Shen, Tian Fang, and Long Quan.
\newblock Recurrent mvsnet for high-resolution multi-view stereo depth
  inference.
\newblock In {\em Proceedings of the IEEE/CVF Conference on Computer Vision and
  Pattern Recognition}, pages 5525--5534, 2019.

\bibitem{yao2020blendedmvs}
Yao Yao, Zixin Luo, Shiwei Li, Jingyang Zhang, Yufan Ren, Lei Zhou, Tian Fang,
  and Long Quan.
\newblock Blendedmvs: A large-scale dataset for generalized multi-view stereo
  networks.
\newblock In {\em Proceedings of the IEEE/CVF Conference on Computer Vision and
  Pattern Recognition}, pages 1790--1799, 2020.

\bibitem{yi2020pyramid}
Hongwei Yi, Zizhuang Wei, Mingyu Ding, Runze Zhang, Yisong Chen, Guoping Wang,
  and Yu-Wing Tai.
\newblock Pyramid multi-view stereo net with self-adaptive view aggregation.
\newblock In {\em European Conference on Computer Vision}, pages 766--782.
  Springer, 2020.

\bibitem{yu2021attention}
Anzhu Yu, Wenyue Guo, Bing Liu, Xin Chen, Xin Wang, Xuefeng Cao, and Bingchuan
  Jiang.
\newblock Attention aware cost volume pyramid based multi-view stereo network
  for 3d reconstruction.
\newblock {\em ISPRS Journal of Photogrammetry and Remote Sensing},
  175:448--460, 2021.

\bibitem{yu2020fast}
Zehao Yu and Shenghua Gao.
\newblock Fast-mvsnet: Sparse-to-dense multi-view stereo with learned
  propagation and gauss-newton refinement.
\newblock In {\em Proceedings of the IEEE/CVF Conference on Computer Vision and
  Pattern Recognition}, pages 1949--1958, 2020.

\bibitem{zhang2020visibility}
Jingyang Zhang, Yao Yao, Shiwei Li, Zixin Luo, and Tian Fang.
\newblock Visibility-aware multi-view stereo network.
\newblock {\em arXiv preprint arXiv:2008.07928}, 2020.

\bibitem{zhang2021learning}
Jingyang Zhang, Yao Yao, and Long Quan.
\newblock Learning signed distance field for multi-view surface reconstruction.
\newblock In {\em Proceedings of the IEEE/CVF International Conference on
  Computer Vision}, pages 6525--6534, 2021.

\bibitem{zhang2021long}
Xudong Zhang, Yutao Hu, Haochen Wang, Xianbin Cao, and Baochang Zhang.
\newblock Long-range attention network for multi-view stereo.
\newblock In {\em Proceedings of the IEEE/CVF Winter Conference on Applications
  of Computer Vision}, pages 3782--3791, 2021.

\bibitem{zhao2018triangle}
Yawei Zhao, Kai Xu, En Zhu, Xinwang Liu, Xinzhong Zhu, and Jianping Yin.
\newblock Triangle lasso for simultaneous clustering and optimization in graph
  datasets.
\newblock {\em IEEE Transactions on Knowledge and Data Engineering},
  31(8):1610--1623, 2018.

\bibitem{zheng2021deep}
Zerong Zheng, Tao Yu, Qionghai Dai, and Yebin Liu.
\newblock Deep implicit templates for 3d shape representation.
\newblock In {\em Proceedings of the IEEE/CVF Conference on Computer Vision and
  Pattern Recognition}, pages 1429--1439, 2021.

\end{thebibliography}
}

\end{document}